\documentclass{article}

\usepackage{iclr2027_conference,times}
\usepackage{amsmath,amssymb,mathtools}
\usepackage{booktabs}
\usepackage{graphicx}
\usepackage{placeins}
\usepackage{xcolor}
\usepackage[appendix=append,bibliography=common,repeqn=independent]{apxproof}
\usepackage{microtype}
\usepackage{hyperref}
\usepackage{url}
\hypersetup{colorlinks=true,linkcolor=blue,citecolor=blue,urlcolor=blue}
\usepackage{subcaption}
\usepackage{bbm}

\title{Variance-Optimal Off-Policy Evaluation\\
with Conjunct Effect Modeling}

\author{Nicolò Felicioni \\
Spotify \\
\And
Michael Benigni \\
Politecnico di Milano \\
\And
Maurizio Ferrari Dacrema \\
Politecnico di Milano \\
\And
Paolo Cremonesi \\
Politecnico di Milano \\
}

\newcommand{\cA}{\mathcal A}
\newcommand{\cC}{\mathcal C}
\newcommand{\D}{\mathcal D}
\newcommand{\E}{\mathbb E}
\newcommand{\Ezero}{\mathbb E_0}
\newcommand{\Vhat}{\widehat V}
\newcommand{\one}{\mathbb I}
\newcommand{\wa}{w_{\mathrm a}}
\newcommand{\wc}{w_{\mathrm c}}
\newcommand{\eqnref}[1]{Eq.~\eqref{#1}}
\newcommand{\method}{\textnormal{\textsc{Variance~Optimal-CEM}}}
\newcommand{\vocem}{\textnormal{\textsc{VOCEM}}}

\DeclareMathOperator{\Var}{Var}
\DeclareMathOperator{\Bias}{Bias}

\theoremstyle{plain}
\newtheoremrep{theorem}{Theorem}[section]
\newtheoremrep{proposition}[theorem]{Proposition}
\newtheoremrep{lemma}[theorem]{Lemma}
\newtheorem{assumption}{Assumption}

\iclrfinalcopy
\begin{document}

\maketitle
\lhead{Preprint}

\begin{abstract}
Off-policy evaluation (OPE) for contextual bandit policies becomes challenging when action-level importance weighting incurs excessive variance.
Doubly robust (DR) estimation remains unbiased under common support but retains these high-variance action-level weights. A prior estimator, Off-policy evaluation with Conjunct Effect Model (OffCEM), replaces them with more stable cluster-level weights, at the cost of relying on local correctness of the reward model.
In this paper, we show that, under the assumptions required by DR and OffCEM, there exists an unbiased family of estimators that interpolates between OffCEM and DR. Building on this result, we propose the \emph{Variance Optimal-CEM} (\vocem{}) estimator, which selects the interpolation coefficient to minimize variance. We derive the population-optimal coefficient in closed form and show that the resulting estimator has variance no larger than either endpoint, OffCEM or DR.
Experiments in controlled synthetic settings and on two large-action benchmarks show that \vocem{} improves upon both endpoints in all 23 evaluated conditions, exhibiting greater stability and empirical robustness.
\end{abstract}

\section{Introduction}
\label{sec:introduction}

Modern recommender systems~\citep{gilotte, saito-johachins-21}, ranking services~\citep{DBLP:conf/www/LiCKG15}, and personalized decision tools
routinely collect contextual bandit feedback~\citep{cmab1,cmab2}.  Each logged interaction records
a context, the action selected by a logging policy, and the resulting reward,
while rewards for the unchosen actions remain unobserved.  \textit{Off-policy
evaluation} (OPE) is the problem of using these data to estimate the value of a different target policy
without deploying it. This would make it possible to evaluate a policy before
an online experiment, but it is a difficult problem to solve.

The standard estimators expose this difficulty at different points.  The
\emph{inverse propensity scoring (IPS)} estimator~\citep{horvitz1952generalization} multiplies every observed reward by an
action-level importance ratio and is unbiased
under common support. However, this action-level importance ratio (which is the ratio between the target policy and logging policy evaluated on that particular action) may explode in many real use cases (say, if the logging policy is very small for some actions), and makes this estimator very unstable due to high variance. The \emph{doubly robust (DR)}
estimator~\citep{dudik2014doubly} uses a reward regression as a control variate and
importance weights only its residual. While this represents an improvement with respect to IPS, the residual correction still contains the same action ratio.
The \emph{Off-policy evaluation with Conjunct Effect Model} (OffCEM)
estimator~\citep{saito2023offcem} instead corrects residuals at the level of
action clusters.  Aggregating policy mass within a cluster makes its importance
ratio more stable, while a reward model represents differences
among actions in that cluster.  They assume local correctness, i.e., the model need only
preserve these within-cluster reward differences rather than recover the
absolute reward.



OffCEM thus appears to take a different approach from DR, replacing
action-level residual weighting with a coarser cluster-level correction.
In this paper, instead, we show that these two estimators are endpoints of a
common family. A single scalar parameter interpolates between their
residual weights, recovering OffCEM at one endpoint and DR at the other.
Under action-level common support and local correctness, every fixed
estimator in this family is unbiased.

This shared unbiasedness gives a natural way to choose among the
estimators in the family. Mean-squared error decomposes into squared
bias and variance; since the bias is zero throughout the family,
minimizing mean-squared error reduces to minimizing variance.
Therefore, in this paper we show that the variance is a quadratic function of the interpolation
parameter and derive its theoretical minimizer in closed form. The resulting
estimator has variance no larger than either OffCEM or DR.
Building on this result, we propose the \method{}
(\vocem{}) estimator, which uses an empirical estimate of the optimal
coefficient to select a member of this family. We then evaluate this practical estimator in controlled
synthetic and real-world experiments. \vocem{} has lower MSE than
OffCEM in all 15 synthetic and eight real-data conditions, 
while the fixed DR endpoint
is often unstable. 

\section{Background}
\label{sec:background}

We work in the contextual bandit setting with a measurable context space
$\mathcal X$, a finite action set $\cA$, and a reward space
$\mathcal R\subseteq\mathbb R$.  Uppercase letters denote random variables
and lowercase letters their realizations.  A policy is a conditional
distribution over $\cA$ given $x\in\mathcal X$; $\pi_0$ denotes the logging
policy and $\pi$ the target policy.  Under $\pi_0$, one observation
$Z=(X,A,R)\in
\mathcal X\times\cA\times\mathcal R$ is generated according to
\begin{equation}
  X\sim p, \qquad A\sim\pi_0(\cdot\mid X), \qquad
  R\sim p(\cdot\mid X,A),
  \label{eq:logging-law}
\end{equation}
where $p$ and $p(\cdot\mid x,a)$ are the context and conditional reward
distributions.  Both are policy-invariant; only the conditional action
distribution changes.  The log
$\D=\{Z_i=(X_i,A_i,R_i)\}_{i=1}^n$ consists of i.i.d. draws from
\eqnref{eq:logging-law}.  We assume that $\pi_0(A_i\mid X_i)$ is known and
that $\pi(a\mid x)$ can be evaluated.

The conditional mean reward $q(x,a)$ and the value $V(\pi)$ of the target policy are
\begin{equation}
  q(x,a):=\E[R\mid X=x,A=a],
  \qquad
  V(\pi)
  :=\E_{p(x)\pi(a\mid x)p(r\mid x,a)}[R]
  =\E_{p(x)\pi(a\mid x)}[q(X,A)].
  \label{eq:policy-value}
\end{equation}
Thus OPE asks for an expectation under the target action distribution using
observations drawn under the logging action distribution.  To make the two
laws unambiguous, for every integrable function $h$ we write
\begin{equation}
  \Ezero[h(Z)]
  :=\E_{p(x)\pi_0(a\mid x)p(r\mid x,a)}[h(X,A,R)].
  \label{eq:logging-expectation}
\end{equation}
Notice that $\Ezero$ always integrates actions against $\pi_0$, whereas
the first expectation in \eqnref{eq:policy-value} integrates them against
$\pi$.

\paragraph{Doubly Robust}
The \emph{doubly robust (DR)} estimator
\citep{dudik2014doubly} approaches OPE by combining reward modeling with importance
weighting.  Let
$\widehat q:\mathcal X\times\cA\to\mathbb R$ be a reward model, i.e., a fitted approximation to
$q$, and define
\begin{equation}
  \widehat q(x,\pi)
  :=\sum_{a\in\cA}\pi(a\mid x)\widehat q(x,a).
  \label{eq:qhat-policy}
\end{equation}
The term $\widehat q(x,\pi)$ is the prediction of the expected reward
under the target policy, known as the direct method (DM) estimator \citep{beygelzimer2009offset}.  To correct its regression error, DR additionally uses action-level importance weighting of the logged
rewards,
requiring the following support condition.

\begin{assumption}[Action-level common support]
\label{ass:action-support}
For $p$-almost every $x$ and every $a\in\cA$,
$\pi(a\mid x)>0$ implies $\pi_0(a\mid x)>0$.
\end{assumption}

Under Assumption~\ref{ass:action-support}, define
\begin{equation}
  \wa(x,a):=\frac{\pi(a\mid x)}{\pi_0(a\mid x)}.
  \label{eq:action-weight}
\end{equation}
For reference, averaging the raw reward weighted by $\wa$ yields the usual \textit{Inverse
Propensity Scoring} (IPS) estimator \citep{horvitz1952generalization}. DR represents an improvement of IPS. Define the model residual $e$
and the pointwise regression error $\Delta(x,a)$:
\begin{equation}
  \begin{aligned}
    e&:=R-\widehat q(X,A),
    \qquad
    \Delta(x,a):=q(x,a)-\widehat q(x,a).
  \end{aligned}
  \label{eq:regression-residual}
\end{equation}
DR applies the weight only to the
reward model residual, correcting the direct estimate,
\begin{equation}
  \Vhat_{\mathrm{DR}}(\pi)
  :=\frac{1}{n}\sum_{i=1}^n
  \left[
    \widehat q(X_i,\pi)
    +\wa(X_i,A_i)e_i
  \right].
  \label{eq:dr}
\end{equation}
For fixed $\widehat q$, the same reweighting calculation gives
\begin{equation}
  \Ezero[\wa(X,A)e]
  =\E_{p(x)\pi(a\mid x)}[\Delta(X,A)],
  \qquad
  \Ezero[\widehat q(X,\pi)+\wa(X,A)e]=V(\pi).
  \label{eq:dr-unbiasedness}
\end{equation}
Thus the weighted residual corrects the bias of the direct estimate.  When
$\widehat q$ is accurate, its residuals can be much less variable than the
raw rewards, which often makes DR more stable than IPS.  The correction,
however, still contains the action-level weight $\wa(X,A)$ and can be
unstable for actions with low logging probability (the ratio would explode).


\section{Action clusters and OffCEM}
\label{sec:offcem}

DR uses ratios at the action-level. In many applications, however, actions come with a useful
coarser organization. This additional information can be used for estimation. Let $\cC$ be a finite set of action clusters and let
$\phi:\mathcal X\times\cA\to\cC$ be a fixed, possibly context-dependent
clustering map.  We write $C=\phi(X,A)$.  A policy induces a distribution over
clusters by aggregating the probabilities of its constituent actions:
\begin{equation}
  \pi(c\mid x)
  :=\sum_{a\in\cA}\one\{\phi(x,a)=c\}\pi(a\mid x),
  \qquad
  \pi_0(c\mid x)
  :=\sum_{a\in\cA}\one\{\phi(x,a)=c\}\pi_0(a\mid x),
  \label{eq:cluster-policies}
\end{equation}
Assumption~\ref{ass:action-support} implies cluster-level support: if
$\pi(c\mid x)>0$, then $\pi_0(c\mid x)>0$.  We can therefore define the
cluster importance weight
\begin{equation}
  \wc(x,c):=\frac{\pi(c\mid x)}{\pi_0(c\mid x)}.
  \label{eq:cluster-weight}
\end{equation}
Because policy mass is pooled across every action in a cluster, these weights
can be far less variable than $\wa(x,a)$.  
The \emph{Off-policy evaluation with Conjunct Effect Model} (OffCEM)
estimator~\citep{saito2023offcem} retains the variance benefit of coarse
importance weights without requiring actions in the same cluster to be
equivalent.  Its starting point is the conjunct effect model, which organizes
the conditional mean reward into a cluster effect and a residual action
effect:
\begin{equation}
  q(x,a)
  =\underbrace{\mu(x,\phi(x,a))}_{\text{cluster effect}}
   +\underbrace{\rho(x,a)}_{\text{residual action effect}}.
  \label{eq:conjunct-effect-model}
\end{equation}
For example, in movie recommendation, $\mu$ can describe a user's preference
for a genre, while $\rho$ distinguishes individual movies within that genre.
OffCEM estimates the change in the cluster effect by importance weighting and
uses the reward model $\widehat q$ to represent the remaining action-level
variation.  This gives the estimator
\begin{equation}
  \Vhat_{\mathrm{OffCEM}}(\pi)
  :=\frac{1}{n}\sum_{i=1}^n
  \left[
    \widehat q(X_i,\pi)+\wc(X_i,\phi(X_i,A_i))e_i
  \right].
  \label{eq:offcem}
\end{equation}
The form resembles DR, but the correction has a different resolution.  The
term $\widehat q(X,\pi)$ averages the fitted rewards under the target policy,
including its choice of actions within each cluster.  The residual is then
reweighted only enough to correct how often the target and logging policies
visit each cluster.  OffCEM thereby avoids multiplying residuals by the
potentially extreme action ratio $\wa(X,A)$.

Since $\wc(x,c)$ is identical for all actions in cluster $c$, it cannot
correct misspecified within-cluster reward differences.  This motivates the
following local requirement.
\begin{assumption}[Local correctness]
\label{ass:local-correctness}
There exists a measurable $b:\mathcal X\times\cC\to\mathbb R$ such that
\begin{equation}
  \Delta(x,a)=b(x,\phi(x,a))
  \qquad\text{for $p$-almost every $x$ and every $a\in\cA$.}
  \label{eq:local-correctness}
\end{equation}
\end{assumption}
Assumption~\ref{ass:local-correctness} permits a shared additive error
$b(x,c)$ because cluster weighting can correct that offset, but it requires
$\widehat q$ to preserve within-cluster reward contrasts.  Indeed, for two
actions $a$ and $a'$ in the same cluster $c$,
\begin{equation}
  q(x,a)-q(x,a')
  =\widehat q(x,a)-\widehat q(x,a').
  \label{eq:within-cluster-differences}
\end{equation}
The additive error may otherwise vary freely across contexts and clusters.

It remains to explain why a cluster-specific error does not bias OffCEM.
Here $\Bias(\Vhat(\pi)):=\E[\Vhat(\pi)]-V(\pi)$.
Subtracting the DR identity in \eqnref{eq:dr-unbiasedness} from the expectation
of \eqnref{eq:offcem}, and then conditioning on $(X,A)$, gives
\begin{equation}
  \Bias(\Vhat_{\mathrm{OffCEM}}(\pi))
  =\Ezero\!\left[
    \bigl(\wc(X,C)-\wa(X,A)\bigr)e
  \right]
  =\Ezero\!\left[
    \bigl(\wc(X,C)-\wa(X,A)\bigr)\Delta(X,A)
  \right].
  \label{eq:offcem-bias}
\end{equation}
The next identity shows that the action ratio averages to the cluster ratio
once $X$ and the cluster $\phi(X,A)$ are fixed.
\begin{lemmarep}[Cluster projection]
\label{lem:cluster-projection}
Under Assumption~\ref{ass:action-support},
\begin{equation}
  \Ezero[\wa(X,A)\mid X,\phi(X,A)]
  =\wc(X,\phi(X,A)).
  \label{eq:projection-identity}
\end{equation}
Consequently, define
\[
  d(X,A):=\wa(X,A)-\wc(X,\phi(X,A)).
\]
For every measurable $h$ such that
$d(X,A)h(X,\phi(X,A))$ is integrable,
\begin{equation}
  \Ezero[d(X,A)h(X,\phi(X,A))]=0.
  \label{eq:cluster-orthogonality}
\end{equation}
\end{lemmarep}
\begin{proof}
Fix $(x,c)$ with $\pi_0(c\mid x)>0$.  The within-cluster logging policy is
\[
  \pi_0(a\mid x,c)
  =\frac{\pi_0(a\mid x)\one\{\phi(x,a)=c\}}{\pi_0(c\mid x)}.
\]
Expanding the conditional expectation and using common support gives
\begin{align*}
  &\Ezero[\wa(X,A)\mid X=x,\phi(x,A)=c] \\
  &=\sum_{a\in\cA}\pi_0(a\mid x,c)
    \frac{\pi(a\mid x)}{\pi_0(a\mid x)} \\
  &=\frac{\sum_{a:\phi(x,a)=c}\pi(a\mid x)}{\pi_0(c\mid x)}
    \\
  &=\frac{\pi(c\mid x)}{\pi_0(c\mid x)}
   =\wc(x,c).
\end{align*}
Here common support ensures that any action with $\pi_0(a\mid x)=0$ also
has $\pi(a\mid x)=0$, so omitting such actions from the conditional sum
does not change its numerator.

We now prove the final claim explicitly.  Conditional on
$X=x$ and $\phi(x,A)=c$, the action $A$ can still vary among the actions in
cluster $c$.  However, $h(x,c)$ and $\wc(x,c)$ are fixed.  Therefore,
they can be taken outside the sum over actions.  By contrast, $\wa(x,a)$,
and hence $d(x,a)$, generally varies with the particular action $a$ and must
remain inside that sum:
\begin{align*}
  &\Ezero[
    d(X,A)h(X,\phi(X,A))
    \mid X=x,\phi(x,A)=c
  ] \\
  &=\sum_{a:\phi(x,a)=c}
    \frac{\pi_0(a\mid x)}{\pi_0(c\mid x)}
    \left\{
      \frac{\pi(a\mid x)}{\pi_0(a\mid x)}
      -\frac{\pi(c\mid x)}{\pi_0(c\mid x)}
    \right\}h(x,c) \\
  &=h(x,c)\left\{
    \frac{1}{\pi_0(c\mid x)}
      \sum_{a:\phi(x,a)=c}\pi(a\mid x)
    -\frac{\pi(c\mid x)}{\pi_0(c\mid x)^2}
      \sum_{a:\phi(x,a)=c}\pi_0(a\mid x)
  \right\} \\
  &=h(x,c)\left\{
    \frac{\pi(c\mid x)}{\pi_0(c\mid x)}
    -\frac{\pi(c\mid x)}{\pi_0(c\mid x)^2}
      \pi_0(c\mid x)
  \right\} \\
  &=h(x,c)\left\{
    \frac{\pi(c\mid x)}{\pi_0(c\mid x)}
    -\frac{\pi(c\mid x)}{\pi_0(c\mid x)}
  \right\}
  =0.
\end{align*}
Finally, the law of iterated expectation gives
\begin{align*}
  &\Ezero[d(X,A)h(X,\phi(X,A))] \\
  &=\Ezero\!\left[
    \Ezero\!\left[
      d(X,A)h(X,\phi(X,A))
      \mid X,\phi(X,A)
    \right]
  \right] \\
  &=\Ezero[0]
  =0.
\end{align*}
\end{proof}

Under local correctness, $\Delta(X,A)=b(X,\phi(X,A))$, so
\eqnref{eq:offcem-bias} is zero by
Lemma~\ref{lem:cluster-projection}.  This is the mechanism behind OffCEM's
unbiasedness: the regression model supplies the within-cluster reward
contrasts, while the cluster importance weight corrects any remaining
cluster-level offset.

Throughout the analysis, $\pi$, $\pi_0$, $\phi$, and $\widehat q$ are fixed.
In particular, $\widehat q$ may be trained on an independent sample, and all
claims can then be read conditionally on that sample.  All first moments
used in the analysis are assumed finite.

\section{Variance Optimal-CEM}
\label{sec:vocem}

DR and OffCEM appear to take different approaches to off-policy evaluation.
In this section, we show that they belong to a more general family of
estimators. A single parameter ranging from $0$ to $1$ traces a path through
this family, with OffCEM and DR as its endpoints.
Under the combined assumptions required by these two estimators, every fixed
member of this family is unbiased. We therefore derive the population
coefficient that minimizes variance along the path and show that the
resulting oracle has variance no larger than either endpoint. We then
estimate this coefficient from the same logging sample to obtain the
\method{} (\vocem{}) estimator.

\subsection{An unbiased path from OffCEM to DR}

Here we introduce a novel family of estimators that includes OffCEM and DR as
special cases. The two estimators share the same direct term; their residual
corrections use the cluster ratio $\wc$ and the action ratio $\wa$, respectively.
For $\lambda\in[0,1]$, define the interpolated weight
\begin{equation}
  \omega_\lambda(x,a)
  :=\wc(x,\phi(x,a))+\lambda d(x,a)
  =(1-\lambda)\wc(x,\phi(x,a))+\lambda\wa(x,a),
  \label{eq:interpolated-weight}
\end{equation}
and the corresponding estimator
\begin{equation}
  \Vhat_\lambda(\pi)
  :=\frac{1}{n}\sum_{i=1}^n
  \left[
    \widehat q(X_i,\pi)+\omega_\lambda(X_i,A_i)e_i
  \right].
  \label{eq:interpolated-estimator}
\end{equation}
The endpoints $\lambda=0$ and $\lambda=1$ recover OffCEM and DR, respectively,
using the same log and reward model. The following proposition shows that
every member is unbiased under the same assumptions required by DR and OffCEM.

\begin{propositionrep}[Exact bias path]
\label{prop:bias-path}
Under Assumption~\ref{ass:action-support}, every fixed
$\lambda\in[0,1]$ satisfies
\begin{equation}
  \Bias(\Vhat_\lambda(\pi))
  =(1-\lambda)\Bias(\Vhat_{\mathrm{OffCEM}}(\pi))
  =(1-\lambda)\Ezero[
    \bigl(\wc(X,C)-\wa(X,A)\bigr)\Delta
  ].
  \label{eq:exact-bias-path}
\end{equation}
If Assumption~\ref{ass:local-correctness} also holds, then
\begin{equation}
  \Bias(\Vhat_\lambda(\pi))=0.
  \label{eq:unbiased-path}
\end{equation}
\end{propositionrep}
\begin{proof}
The estimator mean and target value admit the common decomposition
\begin{align*}
  \E[\Vhat_\lambda(\pi)]
  &=\E[\widehat q(X,\pi)]+\Ezero[\omega_\lambda(X,A)\Delta],\\
  V(\pi)
  &=\E[\widehat q(X,\pi)]+\Ezero[\wa(X,A)\Delta],
\end{align*}
where the second line changes measure from $\pi$ to $\pi_0$.  Hence
\[
  \Bias(\Vhat_\lambda(\pi))
  =\Ezero[\bigl(\omega_\lambda(X,A)-\wa(X,A)\bigr)\Delta]
  =(1-\lambda)\Ezero[
    \bigl(\wc(X,C)-\wa(X,A)\bigr)\Delta
  ].
\]
If local correctness holds, $\Delta=b(X,C)$; the last expectation is zero
by Lemma~\ref{lem:cluster-projection}.
\end{proof}


\subsection{The variance-optimal coefficient and its empirical estimate}

In OPE, the typical way to evaluate the quality of an estimator is via its mean-squared error (MSE). For every fixed $\lambda\in[0,1]$, the MSE is the sum of
squared bias and variance:
\[
  \operatorname{MSE}(\Vhat_\lambda(\pi))
  =\Bias(\Vhat_\lambda(\pi))^2+\Var(\Vhat_\lambda(\pi)).
\]
Under the same assumptions required by OffCEM and DR,
\eqnref{eq:unbiased-path} gives
\[
  \Bias(\Vhat_\lambda(\pi))=0
  \quad\Longrightarrow\quad
  \operatorname{MSE}(\Vhat_\lambda(\pi))=\Var(\Vhat_\lambda(\pi)).
\]
This means that, in order to minimize the MSE, we need to choose among the estimators on the path therefore reduces to minimizing
$\Var(\Vhat_\lambda(\pi))$ over $\lambda\in[0,1]$.
We show how to do so in the following theorem. Before introducing the theorem, we need to define the following auxiliary population quantities:
\begin{equation}
  A:=\Ezero[d^2e^2],
  \qquad
  B:=\Ezero[\wc(X,C)d e^2].
  \label{eq:oracle-quantities}
\end{equation}
These quantities isolate the dependence of the variance on $\lambda$: as the
proof below shows, its $\lambda$-dependent part is
$A\lambda^2+2B\lambda$.  The following theorem gives the minimizing
coefficient and compares its variance with the OffCEM and DR endpoints.

\begin{theoremrep}[Oracle coefficient and variance dominance]
\label{thm:oracle}
Suppose Assumptions~\ref{ass:action-support} and
\ref{ass:local-correctness} hold, the OffCEM and DR summands have finite
second moments, $A$ and $B$ are defined as in \eqnref{eq:oracle-quantities}, and $B$ is finite.  If $A>0$, the coefficient minimizing
$\Var(\Vhat_\lambda(\pi))$ over $\lambda\in[0,1]$ is
\begin{equation}
  \lambda^\star
  :=\Pi_{[0,1]}\!\left(-\frac{B}{A}\right),
  \qquad
  \Pi_{[0,1]}(u):=\min\{1,\max\{0,u\}\}.
  \label{eq:oracle-coefficient}
\end{equation}
If $A=0$, all estimators on the path agree almost surely and we set
$\lambda^\star=0$.  In either case,
\begin{equation}
  \Var(\Vhat_{\lambda^\star}(\pi))
  \leq
  \min\!\left\{
    \Var(\Vhat_{\mathrm{OffCEM}}(\pi)),
    \Var(\Vhat_{\mathrm{DR}}(\pi))
  \right\}.
  \label{eq:variance-dominance}
\end{equation}
The inequality is strict when $A>0$ and $-B/A\in(0,1)$.
\end{theoremrep}
\begin{proof}
Recall the notation and scalar moments below; $\Ezero$ denotes expectation
under the logging law:
\[
  \begin{gathered}
    C=\phi(X,A),\qquad e=R-\widehat q(X,A),\qquad
    d=\wa(X,A)-\wc(X,C),\\
    A=\Ezero[d^2e^2],\qquad B=\Ezero[\wc(X,C)de^2].
  \end{gathered}
\]
Let $Y_\lambda:=\widehat q(X,\pi)+(\wc(X,C)+\lambda d)e$, and let
$Y_{\lambda,i}$ denote its value at $Z_i$.  Then
$\Vhat_\lambda(\pi)=\frac{1}{n}\sum_{i=1}^n Y_{\lambda,i}$.
Since $Y_0,Y_1$ are square-integrable, so are $de=Y_1-Y_0$ and every
$Y_\lambda$.  Moreover,
$\widehat q(X,\pi)de=Y_0de-\wc(X,C)de^2$ is integrable by
Cauchy--Schwarz and the finiteness of $B$.
Proposition~\ref{prop:bias-path} gives
$\Ezero[Y_\lambda]=V(\pi)$ for every fixed $\lambda$.
The cross-term between $\widehat q(X,\pi)$ and $\lambda de$ vanishes:
\[
  \Ezero[\widehat q(X,\pi)de]
  =\Ezero[\widehat q(X,\pi)d\Delta]=0,
\]
where local correctness makes
$\widehat q(X,\pi)\Delta$ a function of $(X,C)$ and
Lemma~\ref{lem:cluster-projection} applies.  Expanding
$Y_\lambda=Y_0+\lambda de$ therefore gives
\begin{align*}
  n\Var(\Vhat_\lambda(\pi))
  &=\Var(Y_0)+2\lambda\Ezero[Y_0de]+\lambda^2\Ezero[d^2e^2]\\
  &=\Var(Y_0)+2B\lambda+A\lambda^2.
\end{align*}
For $A>0$, this is a strictly convex quadratic with unconstrained minimizer
$-B/A$; projection onto $[0,1]$ gives~\eqnref{eq:oracle-coefficient}.  The
segment contains both endpoints, which proves~\eqnref{eq:variance-dominance};
strict convexity gives strict dominance for an interior minimizer.  If
$A=0$, then $de=0$ almost surely, so every estimator on the path is identical.
\end{proof}

A useful practical feature of $\lambda^\star$ is that it can be estimated
directly from the logging sample.  Once $\pi$, $\pi_0$, $\phi$, and
$\widehat q$ are fixed, $A$ and $B$ are expectations of functions of the
logged observation $Z$; neither the unknown conditional reward variance nor
the true regression error is required.  We therefore replace the expectations
in~\eqnref{eq:oracle-quantities} with sample averages:
\begin{equation}
  \begin{aligned}
    \widehat A
    &:=\frac{1}{n}\sum_{i=1}^n d(X_i,A_i)^2e_i^2,\\
    \widehat B
    &:=\frac{1}{n}\sum_{i=1}^n
      \wc(X_i,\phi(X_i,A_i))d(X_i,A_i)e_i^2,\\
    \widehat\lambda
    &:=
    \begin{cases}
      \Pi_{[0,1]}(-\widehat B/\widehat A),&\widehat A>0,\\
      0,&\widehat A=0.
    \end{cases}
  \end{aligned}
  \label{eq:estimated-coefficient}
\end{equation}
The resulting $\widehat\lambda$ is a plug-in estimate of $\lambda^\star$.
Substituting this coefficient into the path gives \method{}:
\begin{equation}
  \begin{aligned}
    \Vhat_{\mathrm{VOCEM}}(\pi)
    &:=\Vhat_{\widehat\lambda}(\pi)\\
    &=\frac{1}{n}\sum_{i=1}^n
      \left[
        \widehat q(X_i,\pi)
        +\bigl(\wc(X_i,\phi(X_i,A_i))
        +\widehat\lambda d(X_i,A_i)\bigr)e_i
      \right].
  \end{aligned}
  \label{eq:vocem}
\end{equation}




\section{Experiments}
\label{sec:experiments}

We investigate whether the VOCEM estimator improves in terms of MSE over DR and OffCEM, first in a controlled synthetic
environment and then on two real-world benchmarks.\footnote{We share the code for the experiments anonymously at \url{https://anonymous.4open.science/r/vocem/}}  The comparisons answer two questions: whether there is a gain that persists across different noise, sample-size, and
action-space regimes in an ideal condition of perfect reward model, and whether this gain persists when the reward model is not perfect.
To keep the presentation focused, in this section we restrict the comparison to DR, OffCEM, and VOCEM. In Appendix~\ref{app:additional-baselines}, we report results for the same experimental settings including additional baseline OPE estimators.
In every
condition, compared estimators share the same logging data, reward regression,
clustering, and policy probabilities. To analyze the outcomes of the experiments, we report the relative MSE for each estimator i.e., $\E[(\Vhat(\pi)-V(\pi))^2]/V(\pi)^2$, normalized by the relative MSE of OffCEM.
Thus, the OffCEM reference is always one, and smaller values are better.

\subsection{Synthetic data}
\label{sec:synthetic-experiments}

\subsubsection{Setup}
We build on the public OffCEM synthetic data generator and its default
configuration~\citep{saito2023offcem}: we generate 200 distinct users represented
by 10-dimensional contexts, actions have 10 categorical attributes with five
levels each, and up to 50 action clusters are used.  The conditional mean
reward $q(x,a)$ is generated by a cluster component and an action component,
as in the OffCEM construction \citep{saito2023offcem}. Following \citep{saito2023offcem}, we likewise use a softmax logging policy
with inverse temperature $\beta=-0.1$, an $\epsilon$-greedy target policy with
$\epsilon=0.2$, and the default values $n=3{,}000$ and $|\cA|=1{,}000$.

The only deliberate change to the reward generator of \citet{saito2023offcem} is heteroskedasticity i.e., we make the reward variance different across actions. To induce within-cluster heteroskedasticity, we fix an ordering of the actions in each cluster and map their indices linearly onto \([0,1]\). Denoting the resulting value for action \(a\) by \(u(a)\), we set
\begin{equation}
  \Var(R\mid X,A=a)=9\,\kappa^{u(a)}.
  \label{eq:synthetic-noise}
\end{equation}
At $\kappa=1$, this is exactly the constant standard deviation $\Var(R\mid X,A)=3$ used by
\citet{saito2023offcem}. To show the behaviours of \vocem{} in different regimes, we execute the same experiment with varying parameters: (i) $\kappa\in\{1,2,4,8,16\}$; (ii)
$n\in\{500,1000,2000,4000,8000\}$ at $\kappa=16$; and (iii)
$|\cA|\in\{200,400,1000,2000,4000\}$ at $\kappa=16$.  All other
parameters remain at the defaults above. Each condition is
repeated 300 times. Appendix~\ref{app:synthetic-protocol}
specifies the generator, randomization, and sweep controls in full.

The conditional mean $q(x,a)$ is known in this synthetic experiment, and all three estimators use
$\widehat q=q$.  This removes reward model error and isolates the correction
weights: OffCEM uses $\wc(X,C)$, DR uses $\wa(X,A)$, and VOCEM estimates
$\widehat\lambda$ from all $n$ logged observations.

\begin{figure}
  \centering
  \includegraphics[width=\linewidth]{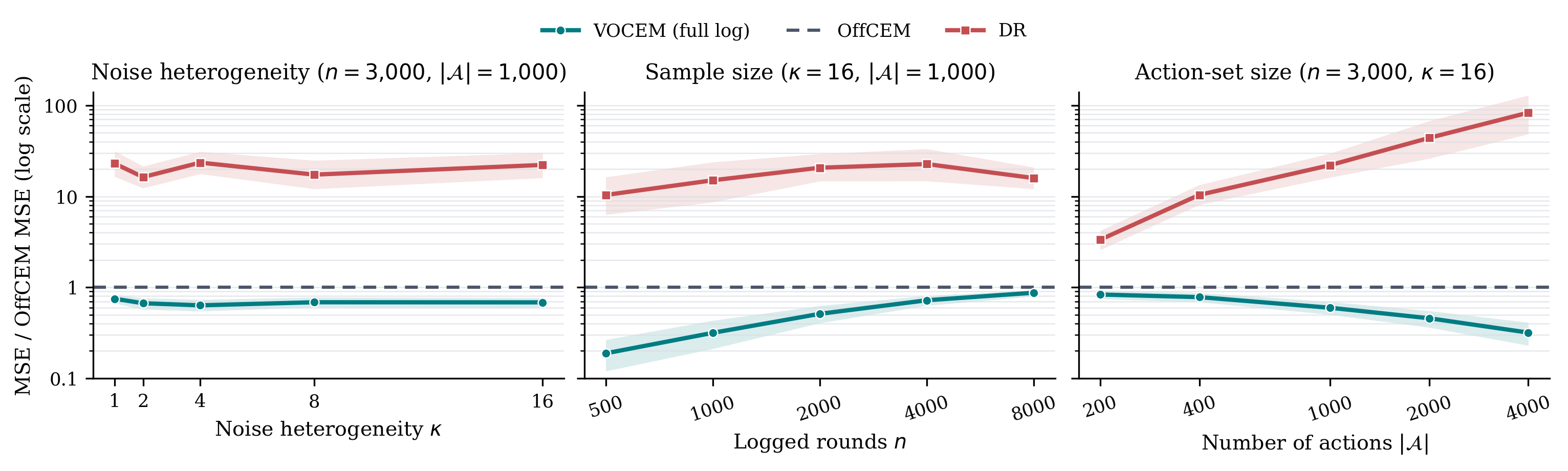}
  \caption{Results on synthetic data.  Points are MSE ratios to OffCEM and
  ribbons are pointwise paired-bootstrap 95\% confidence intervals over 300
  repetitions.  The dashed line is OffCEM.  Only the same-log VOCEM estimator
  is shown; every VOCEM interval lies below one.}
  \label{fig:synthetic-full-sample}
\end{figure}

\subsubsection{Results}
Figure~\ref{fig:synthetic-full-sample} shows a uniform advantage for VOCEM:
its MSE is below that of OffCEM in all 15 conditions, and every pointwise
upper confidence bound is smaller than one.  Across the noise heterogeneity sweep,
the MSE ratio ranges from $0.634$ to $0.749$, a reduction of
$25.1$--$36.6\%$.  The VOCEM gain is present even at $\kappa=1$, where the noise
model reduces to the original OffCEM's homoskedastic setting, and remains substantial at
$\kappa=16$ (MSE ratio of $0.684$).

For the sample size sweep, the MSE ratio increases from $0.188$ at $n=500$ to $0.873$ at $n=8{,}000$, approaching the OffCEM endpoint.  Conversely, the benefit becomes larger with the action space: the
ratio falls from $0.834$ at 200 actions to $0.318$ at 4,000 actions.  In that
same sweep, DR deteriorates from $3.34$ to $83.38$ times the OffCEM MSE,
whereas VOCEM remains below OffCEM throughout.  VOCEM therefore exploits the
action-level benefits without inheriting DR's instability
in the large-action regime.

\begin{figure}
  \centering
  \includegraphics[width=\linewidth]{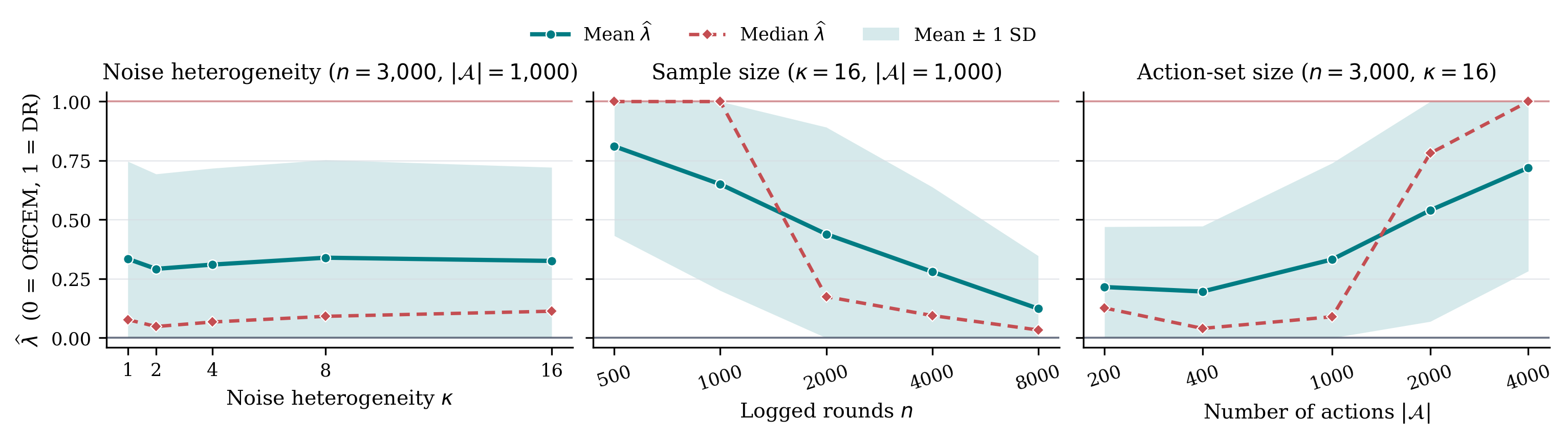}
  \caption{The $\widehat\lambda$ coefficient in the synthetic experiments.  Solid
  curves and dashed curves show the mean and median of $\widehat\lambda$ over
  300 repetitions; ribbons span one standard deviation around the mean,
  clipped to $[0,1]$.  The endpoints $0$ and $1$ recover OffCEM and DR,
  respectively.  The ribbons describe run-to-run variation, not confidence
  intervals for the mean.}
  \label{fig:synthetic-lambda}
\end{figure}

\paragraph{Behavior of $\widehat{\lambda}$ coefficient.}
Figure~\ref{fig:synthetic-lambda} clarifies how VOCEM obtains these gains.
Changing only the noise heterogeneity has little effect on the average $\widehat\lambda$
coefficient: its mean remains between $0.292$ and $0.339$.  Its median is much
closer to the OffCEM endpoint ($0.048$--$0.113$), reflecting a broad, endpoint-heavy
distribution: $25$--$33\%$ of repetitions select $\widehat\lambda=0$, while
$18$--$23\%$ select $\widehat\lambda=1$.

The other two sweeps expose stronger adaptation.  As $n$ increases from 500
to 8,000, the mean $\widehat\lambda$ coefficient falls from $0.811$ to $0.124$, its median falls
from $1$ to $0.033$, and the fraction at the DR endpoint, $\widehat\lambda=1$, falls from $79\%$ to
$1\%$.  This movement toward OffCEM agrees with the diminishing MSE advantage
in the middle panel of Figure~\ref{fig:synthetic-full-sample}.  In contrast,
increasing $|\cA|$ from 200 to 4,000 raises the mean from $0.215$ to $0.720$;
the median reaches $1$, and $70\%$ of repetitions select the DR endpoint in
the largest action space.  This does not imply that the DR estimator is reliable: Figure~\ref{fig:synthetic-full-sample} shows the opposite. Although VOCEM selects the DR endpoint in many repetitions, it shrinks the action-level correction toward OffCEM in the smaller set of repetitions where action-level weighting is unstable. These repetitions with unstable weights account for a large share of DR's MSE, so avoiding its most extreme errors substantially reduces VOCEM's overall MSE.

\FloatBarrier
\subsection{Real-world data}
\label{sec:real-experiments}

\subsubsection{Datasets and protocol}
Following the real-data evaluation of~\citet{saito2023offcem}, we use the
EUR-Lex 4K and Wiki10-31K datasets from the Extreme Classification
Repository~\citep{bhatia2016extreme}.  Each instance consists of a document
$x$ and a set $\mathcal{L}(x)\subseteq\mathcal{A}$ of labels assigned to that
document in the original dataset. We regard a document
and action pair $(x,a)$ as relevant when $a\in\mathcal{L}(x)$.  These
relevance annotations are used to construct the semi-synthetic rewards
described below.

Before constructing the bandit problem, we filter the Wiki10-31K action set
using the rule adopted by \cite{saito2023offcem}.  For each label $a$, we count the number of
documents to which it is assigned across the provided training and test
splits. We retain label $a$ as an action only if it is assigned to more than $9$ documents. This filtering retains 5,501 of the
30,938 Wiki10-31K labels.  We retain all 3,956 labels for EUR-Lex 4K.

\begin{table}[t]
  \caption{Real-world dataset statistics.  ``Evaluated actions'' records the
  action set after the stated frequency filtering.}
  \label{tab:dataset-statistics}
  \centering
  \small
  \begin{tabular}{lrrrr}
    \toprule
    Dataset & Training documents & Test documents
        & Original labels & Retained actions \\
    \midrule
    EUR-Lex 4K  & 15,449 & 3,865 & 3,956  & 3,956 \\
    Wiki10-31K  & 14,146 & 6,616 & 30,938 & 5,501 \\
    \bottomrule
  \end{tabular}
\end{table}

We convert the original relevance annotations into stochastic binary rewards
using the semi-synthetic construction used in \cite{saito2023offcem}.  At dataset construction,
each retained action $a$ is assigned a fixed offset
$\eta_a\in[0,0.2]$, sampled independently and held fixed throughout the
experiment.  Define
\begin{equation}
  q(x,a)
  =
  \begin{cases}
    \sigma(1-\eta_a), & a\in\mathcal{L}(x),\\
    \sigma(\eta_a-1), & a\notin\mathcal{L}(x),
  \end{cases}
\end{equation}
where $\sigma$ is the sigmoid function. We then draw
$
  R\mid X=x,A=a\sim\operatorname{Bernoulli}\,(q(x,a)).
$
The action-specific offset $\eta_a$ controls the separation
between these two probabilities: larger values make relevance less easily
distinguishable from irrelevance.

As in OffCEM,
the logging policy is a softmax over ridge-predicted rewards with
$\beta=30$, the target policy is $\epsilon$-greedy with $\epsilon=0.1$, and actions
are assigned uniformly at random to 100 clusters.  We sample 300 users,
vary the number of logged rounds over
$n\in\{1000,2000,4000,8000\}$, and run 100 paired repetitions.

We implement the reward model as a ridge regression model using as input document features, compressed to 100 components with truncated
SVD. We train the model by five-fold cross-fitting.
All observations from the same user are assigned to the same fold.  For each
fold $k$, we fit the ridge model using observations from users in the other four folds. For every logged observation in fold $k$, we compute the out-of-fold
prediction and the corresponding residual.
After obtaining one out-of-fold residual for every row, we pool all $n$
residuals and estimate a single coefficient $\widehat\lambda$ using
Equation~\eqref{eq:estimated-coefficient}. 

This experiment evaluates the VOCEM estimator with a
learned reward model. 
Appendix~\ref{app:real-protocol} gives the complete preprocessing, policy,
regression, and seed specifications. Appendix \ref{app:deficient-support} reports synthetic experiments under violations of the common action support assumption.

\begin{figure}[!htb]
  \centering
  \includegraphics[width=0.82\linewidth]{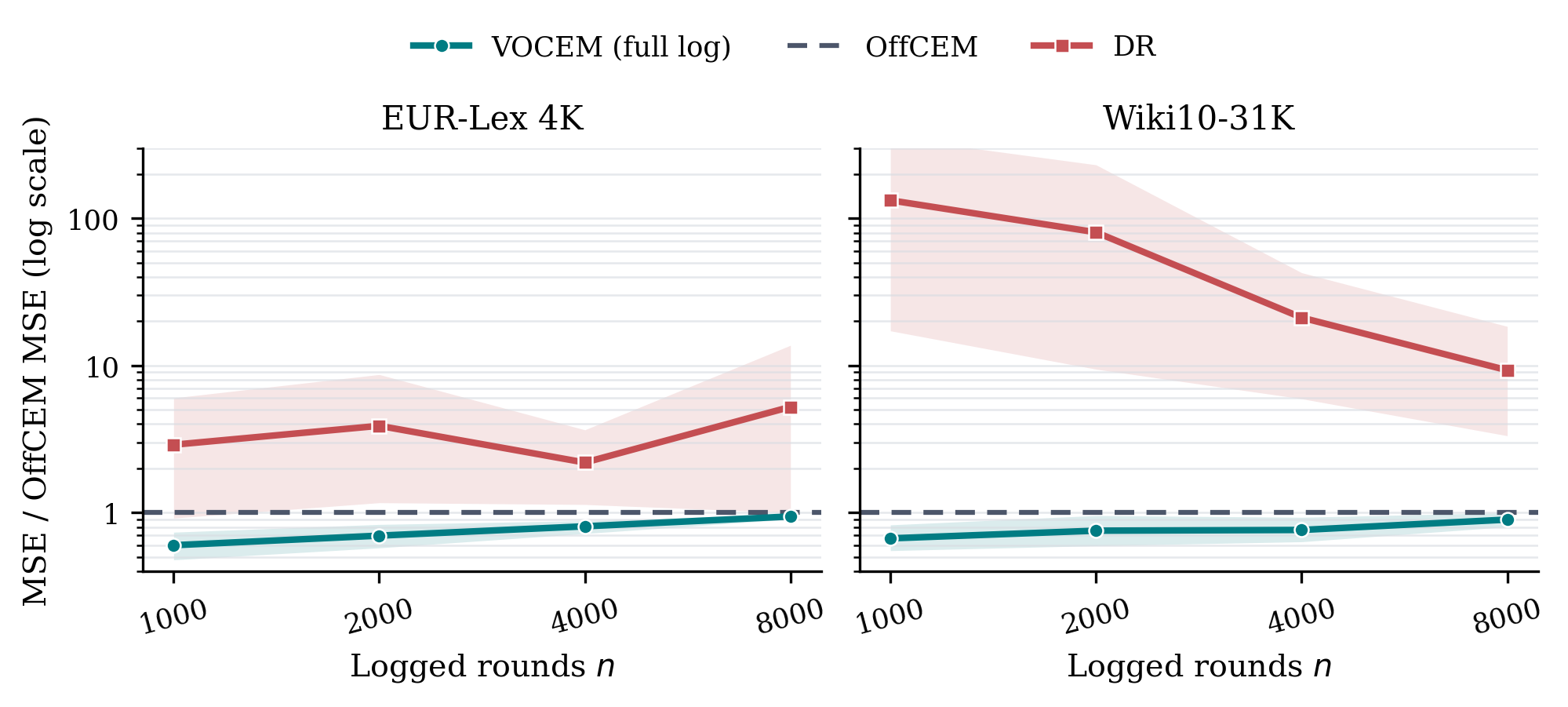}
  \caption{Results on real-data with the five-fold out-of-fold ridge
  reward model.  Points are MSE ratios to OffCEM and ribbons are pointwise
  paired-bootstrap 95\% confidence intervals over 100 repetitions.  The
  dashed line is OffCEM.}
  \label{fig:real-full-sample}
\end{figure}

\subsubsection{Results}
Figure~\ref{fig:real-full-sample} again favors VOCEM in every condition.  On
EUR-Lex 4K, its MSE ratio ranges from $0.601$ at $n=1{,}000$ to $0.942$ at
$n=8{,}000$, corresponding to the two
ends of the sweep.  On Wiki10-31K, the ratios range from $0.669$ to $0.897$.  All eight pointwise upper confidence
bounds remain below one; the closest is $0.999$ for Wiki10-31K at
$n=8{,}000$.  DR is markedly less stable, with MSE between
$2.19$ and $132.94$ times the OffCEM one. These results show that VOCEM can produce gains even with estimated rewards, while
retaining the protection afforded by cluster-level weighting in large action
spaces.

\begin{figure}[!htb]
  \centering
  \includegraphics[width=0.72\linewidth]{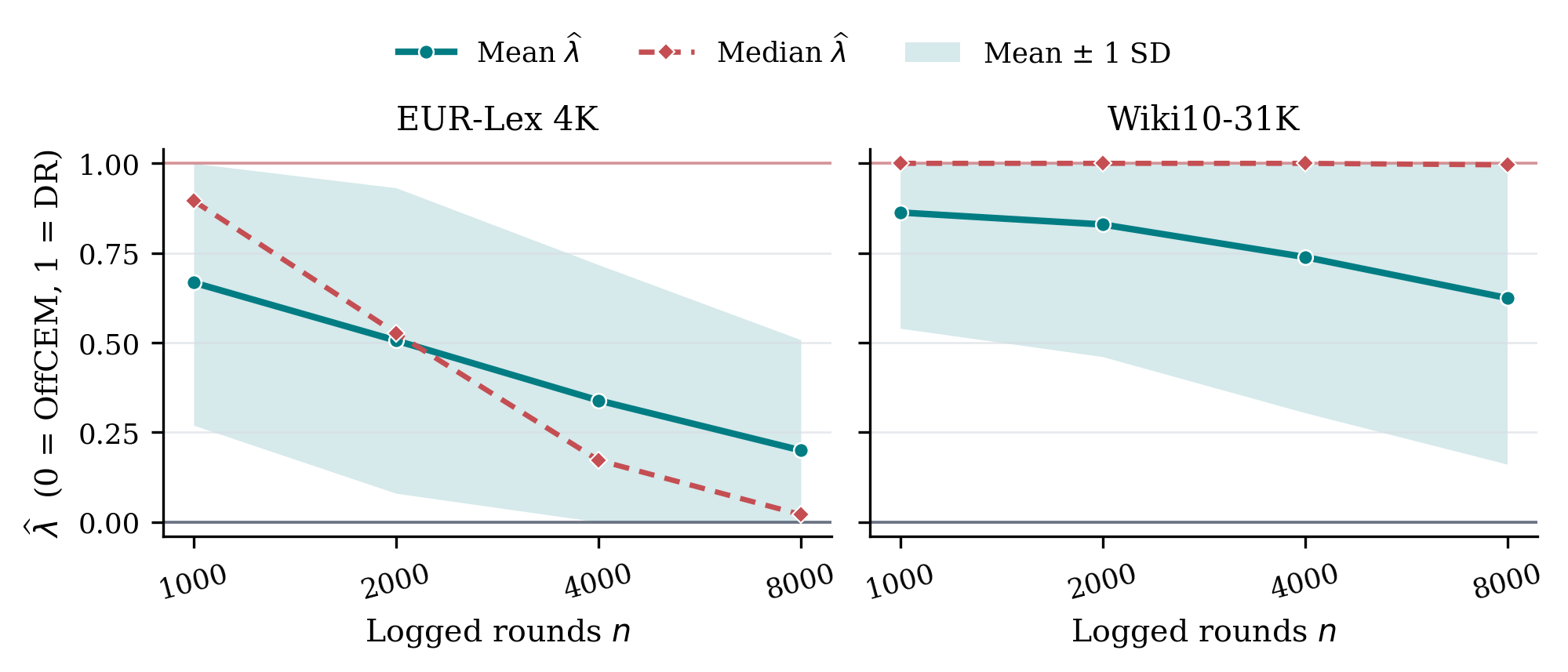}
  \caption{The $\widehat\lambda$ coefficient with learned rewards on EUR-Lex 4K and
  Wiki10-31K.  Solid curves and dashed curves are the mean and median over 100
  repetitions; ribbons span one standard deviation around the mean, clipped
  to $[0,1]$.}
  \label{fig:real-lambda}
\end{figure}

\paragraph{Coefficient behavior.}
The coefficient paths in Figure~\ref{fig:real-lambda} differ substantially
between datasets.  On EUR-Lex 4K, the mean falls from $0.667$ to $0.200$ as
$n$ grows, while the median falls from $0.895$ to $0.020$.  The probability of
the OffCEM endpoint consequently rises from $19\%$ to $48\%$, and no
repetition selects the DR endpoint at $n=8{,}000$.  Wiki10-31K places much
more weight on the action-level correction: the mean decreases more gradually
from $0.863$ to $0.624$, and the median remains essentially at the DR endpoint.
The fraction with $\widehat\lambda=1$ nevertheless declines from $79\%$ to
$46\%$ as the log grows.

The wide ribbons and the separation between means and medians are consequences
of substantial between-log diversity.  They also explain why
the mean coefficient should not be read as a fixed global interpolation.  In
particular, Wiki10-31K often selects $\widehat\lambda=1$, yet applying DR to
every log is extremely unstable in Figure~\ref{fig:real-full-sample}.  VOCEM's
advantage comes from making the endpoint decision separately on each full log,
including backing away from DR on the minority of logs responsible for its
largest errors.

An oracle-regression ablation in
Appendix~\ref{app:oracle-regression-ablation} reaches the same qualitative
conclusion after removing reward-model estimation error.

\nosectionappendix
\begin{toappendix}
\clearpage
\section{Related Work}
\label{app:related-work}

High variance is a central obstacle in off-policy evaluation.  Inverse
propensity scoring (IPS) changes measure from the logging policy to the target
policy and is unbiased under common support, but even a small number of actions
that are likely under the target policy and rare under the logging policy can
produce extreme importance ratios and dominate the estimate~\citep{horvitz1952generalization,ips1,ips2}.
This difficulty has long been recognized in both reinforcement learning and
contextual bandits.  The direct method (DM)~\citep{beygelzimer2009offset} avoids importance weighting by
imputing missing rewards with a reward model, thereby reducing variance at the
cost of bias under model misspecification.  Doubly robust (DR) estimation
combines these approaches by using the reward model as a control variate and
importance weighting only its residual~\citep{dr1,dr2,dudik2014doubly}.  Although accurate reward
predictions can make these residuals much less variable than raw rewards, DR
retains the same action importance ratio as IPS.  Consequently, a rare
action with a large residual can still make DR highly unstable, with the
severity depending on policy overlap and reward model quality.  Work on
debiased OPE studies how nuisance estimation and high-dimensional context
information affect efficiency and uncertainty quantification in recommender
systems~\citep{narita}.  Complementing this frequentist line,
\citet{DBLP:conf/recsys/Benigni25} advocates a Bayesian perspective that
propagates reward-model uncertainty to policy-value estimates
under model misspecification.

Much of the OPE literature can therefore be understood as seeking protection
against extreme importance-weighted observations \citep{DBLP:conf/icml/SakhiAC23,DBLP:conf/icml/SuWSJ19,implexp1,DBLP:conf/aistats/SondhiAD20}.  The simplest approach is
self-normalization, which improves stability by dividing by the empirical
average importance weight~\citep{snips1,snips2}.  More sophisticated IPS
corrections have also been proposed.  \citet{subgaus} introduce a power-mean
transformation that achieves subgaussian concentration under suitable
conditions, while \citet{logsmoothips} consider a class of regularized IPS
estimators and derive an optimal logarithmic smoothing rule.  Model-based
approaches instead exploit a bias--variance tradeoff.  SWITCH relies on the
reward model in high weight regions~\citep{switch}, while shrinkage-based DR
estimators modify the residual importance weights by optimizing a bound on
mean squared error~\citep{dros}.  Marginal density ratio estimation reweights
changes in the outcome distribution rather than the full action policy and can
improve over action ratios~\citep{marginalratio}.

Because the best stabilization strategy depends on overlap, sample size,
reward noise, and the target policy, a parallel literature chooses or combines
estimators from data \citep{voloshin}.  SLOPE selects estimators from an ordered family using
estimated error criteria~\citep{slope}; PAS-IF constructs pseudo evaluation
policies tailored to the target OPE task~\citep{saito23}; and more recent
proposals use cross-validation, reweighted aggregation, or learned automatic
selection~\citep{cief24,nie2024opera,DBLP:journals/corr/abs-2406-18022}.
These methods address the fact that no single conventional estimator is
uniformly best, but they treat the candidate estimators largely as black boxes.
Our work instead derives the variance-optimal estimator analytically within a
particular structured family of estimators.

Large action spaces are an especially important instance of the high-variance
problem \citep{DBLP:conf/aistats/AoualiBRK25,DBLP:journals/corr/abs-2509-03456,DBLP:conf/recsys/ShimizuTKKNS24}.  As policy mass is spread across more actions, individual logging
probabilities can become very small, and the mismatch between logging and
target policies can generate extreme action-level ratios.  At the limit, some
target actions may have no logging support at all.  
Marginalized IPS (MIPS)
uses marginal importance weights when action embeddings provide structure in the action space~\citep{saito2022embeddings}.  Policy convolution~\citep{sachdeva2024off} uses action embeddings to leverage action similarity and
control the bias--variance tradeoff.  Work on
deficient support uses restrictions on the target policy or additional action
information~\citep{def-support}; in particular, side information can support
unbiased estimation under structural assumptions even when ordinary IPS is
biased~\citep{DBLP:conf/nips/FelicioniDRC22}.  These methods show how action
structure can alleviate the general variance and support problems, but their
guarantees depend on the representation or structural assumptions being used.

OffCEM is the most direct precursor to our work.  It replaces the action weight
in the DR residual correction with a more stable cluster weight and uses a
reward model to preserve within-cluster correctness~\citep{saito2023offcem}.  Under local model correctness, this coarser
correction remains unbiased.  VOCEM starts from the observation that,
under common action support and local correctness assumptions, both OffCEM and DR are unbiased. We show that they are
endpoints of a family of unbiased estimators, derive the exact variance along that
family, and obtain the population-optimal coefficient that minimizes the variance in closed form.  The
resulting population estimator has variance no larger than either OffCEM or
DR.  Thus VOCEM addresses excessive variance by adapting the resolution of
importance weighting from clusters toward individual actions, while
preserving the common unbiasedness property.

\section{Detailed Experimental Settings and Additional Results}
\label{app:experimental-details}

Following the organization of the OffCEM experimental
appendix~\citep{saito2023offcem}, we record here the details needed to
reconstruct the data-generating processes and every reported statistic.  We
also identify which choices reproduce the OffCEM benchmark and which are
specific to our implementation.

\subsection{Estimator implementation and reporting}
\label{app:common-protocol}

\paragraph{Full-log score.}
All methods in a comparison use exactly the same log, policy probabilities,
cluster map, and reward predictions.  Let $e_i=R_i-\widehat q(X_i,A_i)$ and
$d_i=\wa(X_i,A_i)-\wc(X_i,C_i)$.  We compute
\begin{equation}
  \widehat A=\frac{1}{n}\sum_{i=1}^n d_i^2e_i^2,
  \qquad
  \widehat B=\frac{1}{n}\sum_{i=1}^n \wc(X_i,C_i)d_ie_i^2,
  \qquad
  \widehat\lambda=\Pi_{[0,1]}(-\widehat B/\widehat A),
  \label{eq:appendix-empirical-lambda}
\end{equation}
and evaluate that coefficient on the same $n$ observations.  Numerically, we
use the equivalent affine identity
\begin{equation}
  \Vhat_{\mathrm{VOCEM}}(\pi)
  =\Vhat_{\mathrm{OffCEM}}(\pi)
   +\widehat\lambda
    (\Vhat_{\mathrm{DR}}(\pi)-\Vhat_{\mathrm{OffCEM}}(\pi)).
  \label{eq:appendix-affine-score}
\end{equation}

\paragraph{Pairing, error metric, and uncertainty.}
For replication $b$, let $V_b$ denote its value under the target policy $\pi$ and
$E_{b,m}=\{(\Vhat_{b,m}(\pi)-V_b)/V_b\}^2$ the relative squared error of method
$m$.  Figures report
\begin{equation}
  \frac{B^{-1}\sum_{b=1}^B E_{b,m}}
       {B^{-1}\sum_{b=1}^B E_{b,\mathrm{OffCEM}}}.
  \label{eq:appendix-mse-ratio}
\end{equation}
We use
$B=300$ for synthetic conditions and $B=100$ for real-data conditions. Every
ribbon is a pointwise percentile interval from 2,000 paired bootstrap
resamples.   The
$\widehat\lambda$ ribbons are different: they show one standard deviation around the mean, clipped only for display to $[0,1]$;
the dashed curve reports the median.  No model or plotting hyperparameter was
chosen by inspecting the reported MSE curves.

\subsection{Synthetic benchmark}
\label{app:synthetic-protocol}

\paragraph{OffCEM-matched environment.}
We retain the public OffCEM generator~\citep{saito2023offcem} and its random
seed 12,345.  It first draws 200 user vectors independently from
$\mathcal N(0,I_{10})$.  Every action receives ten categorical attributes,
each with five possible levels; the attribute distribution
and the cluster assignment of actions are generated exactly by the released
OffCEM code.  The requested number of actions in a cluster is 50.  
The conditional mean has the original decomposition as in \cite{saito2023offcem}
\begin{equation}
  q(x,a)=g(x,\phi(a))+h_{\phi(a)}(x,a),
  \label{eq:appendix-synthetic-mean}
\end{equation}
where $g$ is a polynomial cluster-effect function and $h_c$ is a action-residual linear effect. For each user, the logging and target probabilities are
\begin{equation}
  \pi_0(a\mid x)
  =\frac{\exp\{-0.1q(x,a)\}}
         {\sum_{a'\in\cA}\exp\{-0.1q(x,a')\}},
  \qquad
  \pi(a\mid x)
  =0.8\,\one\!\left\{a=a^\star(x)\right\}
   +\frac{0.2}{|\cA|},
  \label{eq:appendix-synthetic-policies}
\end{equation}
using the same $\beta=-0.1$ and $\epsilon=0.2$ choices as OffCEM, where
$a^\star(x)$ is the greedy action returned by the released policy routine
(including its tie breaking).  A logged round
draws a user uniformly from the 200 fixed users, an action from $\pi_0$, and
then
\begin{equation}
  R=q(X,A)+9\,\kappa^{u(A)}Z,
  \qquad Z\sim\mathcal N(0,1),
  \label{eq:appendix-synthetic-reward}
\end{equation}
where $u(a)$ is a linear projection of a fixed ordering of the actions in each cluster.  Setting
$\kappa=1$ recovers OffCEM's homoskedastic reward with a variance of 3.

Every estimator receives the same
oracle regression $\widehat q=q$, so all differences arise from correction
weights.

\begin{table}
  \caption{Synthetic sweeps.  Only the first column varies within a row.}
  \label{tab:appendix-synthetic-sweeps}
  \centering
  \small
  \begin{tabular}{lll}
    \toprule
    Varied quantity & Values & Quantities held fixed \\
    \midrule
    Heterogeneity $\kappa$ & $1,2,4,8,16$ & $n=3{,}000$, $|\cA|=1{,}000$ \\
    Logged rounds $n$ & $500,1{,}000,2{,}000,4{,}000,8{,}000$
      & $\kappa=16$, $|\cA|=1{,}000$ \\
    Actions $|\cA|$ & $200,400,1{,}000,2{,}000,4{,}000$
      & $n=3{,}000$, $\kappa=16$ \\
    \bottomrule
  \end{tabular}
\end{table}

\subsection{Real-world semi-synthetic benchmark}
\label{app:real-protocol}

\paragraph{Dataset preparation.}
We use the standard train--test splits of EUR-Lex 4K and Wiki10-31K reported in Table~\ref{tab:dataset-statistics}.  Their sparse bag-of-words
representations have respectively 186,104 and 101,938 input coordinates.
Following \citet{saito2023offcem}, the original test split is used only to construct the
logging policy, while evaluation users are drawn from the training split.
We retain all EUR-Lex actions and apply the same OFFCEM filtering to Wiki labels (as detailed in Section \ref{sec:experiments}).  This
leaves 3,956 and 5,501 evaluated actions, respectively.

Our scalable preprocessing differs from the original implementation of OffCEM.
We select 2,000 test documents without replacement, fit a 100-component
ARPACK truncated SVD on their sparse features, and standardize the resulting
coordinates.  The same transformation is then applied to the training
documents.  On those 2,000 reference documents, a
ridge model predicts the conditional reward for every
action,
\begin{equation}
  \widetilde q_{\mathrm b}(x,a)=\alpha_a+z(x)^\top\theta_a,
  \qquad
  \pi_0(a\mid x)
  =\frac{\exp\{30\widetilde q_{\mathrm b}(x,a)\}}
         {\sum_{a'}\exp\{30\widetilde q_{\mathrm b}(x,a')\}}.
  \label{eq:appendix-real-logging}
\end{equation}
The use of a ridge model and inverse temperature $30$
follows OffCEM.

\paragraph{Reward and target policy.}
A dataset-level draw assigns each action
$\eta_a\sim\mathrm{Unif}(0,0.2)$.  If $Y_{xa}$ denotes whether label $a$ is
relevant to document $x$, the Bernoulli mean is
\begin{equation}
  q(x,a)=
  \begin{cases}
    \sigma(1-\eta_a),&Y_{xa}=1,\\
    \sigma(\eta_a-1),&Y_{xa}=0,
  \end{cases}
  \qquad \sigma(t)=(1+e^{-t})^{-1}.
  \label{eq:appendix-real-reward}
\end{equation}
This is the stochastic binary-reward construction of OffCEM.  Let
$a^\star(x)$ maximize this known mean.  Its main-text target-policy setting is
also retained:
\begin{equation}
  \pi(a\mid x)=0.9\,\one\{a=a^\star(x)\}+\frac{0.1}{|\cA|}.
  \label{eq:appendix-real-target}
\end{equation}

For each of 100 replications, we sample 300 distinct training documents as
users, then draw an $8{,}000$-round log by sampling users uniformly with
replacement, actions from~\eqnref{eq:appendix-real-logging}, and Bernoulli
rewards from~\eqnref{eq:appendix-real-reward}.  The four sample sizes are the
nested prefixes $n\in\{1{,}000,2{,}000,4{,}000,8{,}000\}$ of this log.  Each
action is independently assigned uniformly to one of 100 clusters once per
replication, and that partition is shared by every prefix and estimator.
These choices match the two datasets, $\beta=30$, $\epsilon=0.1$, 100 random
clusters, and 100 repetitions used in the main real-data study of OffCEM.

\begin{table}
  \caption{Overlap diagnostics for the constructed real-data policies.  The
  extrema range over all constructed users and replications.  Ratios are not
  clipped.}
  \label{tab:appendix-overlap}
  \centering
  \small
  \begin{tabular}{lrrr}
    \toprule
    Dataset & Evaluated actions & $\min\pi_0(a\mid x)$ & $\max\wa(x,a)$ \\
    \midrule
    EUR-Lex 4K & 3,956 & $7.45\times10^{-9}$ & $4.07\times10^7$ \\
    Wiki10-31K & 5,501 & $1.14\times10^{-14}$ & $1.60\times10^9$ \\
    \bottomrule
  \end{tabular}
\end{table}

The common-support assumption holds by construction, but
Table~\ref{tab:appendix-overlap} shows that practical overlap is weak.  This
explains why the action-weighted DR endpoint can have very high Monte Carlo
MSE even though all probabilities are known exactly.

\paragraph{Learned reward regression.}
We use five-fold group cross-fitting, with the 300 user identifiers as groups:
all appearances of one user are assigned to the same fold.  On the other four
folds, we fit the additive ridge model
\begin{equation}
  \widehat q_{-k}(x,a)
  =\Pi_{[0,1]}\!\left(\alpha_k+z(x)^\top\theta_k+\gamma_{k,a}\right),
  \label{eq:appendix-real-nuisance}
\end{equation}
using squared loss, penalty 1, tolerance $10^{-4}$, and at most 1,000 solver
iterations.  Each held-out row receives exactly one prediction.  We compute
$\widehat q_{-k}(x,\pi)$ by evaluating and clipping the model action by action
and then summing with the known target probabilities; it is not an
approximation based only on the observed action.  After concatenating the five
held-out blocks, all OOF residuals enter the single complete-log coefficient
in~\eqnref{eq:appendix-empirical-lambda}.  Thus reward prediction is OOF at the
user level, while selection of $\widehat\lambda$ deliberately reuses the
complete set of OOF residuals.  This additive ridge nuisance is not the
three-hidden-layer, two-stage neural regression of OffCEM, a distinction that
prevents attributing our learned-regression comparison to that pipeline.

\paragraph{Seeds and target value.}
Feature reduction, action noise, and the behavior regression use seed 123.
Replication $b$ uses seed $123+b$ for EUR-Lex and
$1{,}000{,}123+b$ for Wiki; its cluster map uses that seed plus 500,000.  The
target value $V_b$ is computed exactly.

\subsection{Oracle-regression ablation on real-world data}
\label{app:oracle-regression-ablation}

The main real-data study deliberately uses an estimated reward model.  To
separate its errors from the choice of correction weight, we repeat the same
full-log protocol with the conditional reward mean supplied to every
estimator, $\widehat q=q$.  All other ingredients---datasets, policies,
random 100-cluster partition, sample sizes, paired repetitions, and bootstrap
construction---are unchanged.  This is the real-data analogue of the
oracle-regression design already used throughout the synthetic study, so a
second synthetic ablation would be redundant.

\begin{figure}
  \centering
  \includegraphics[width=0.82\linewidth]{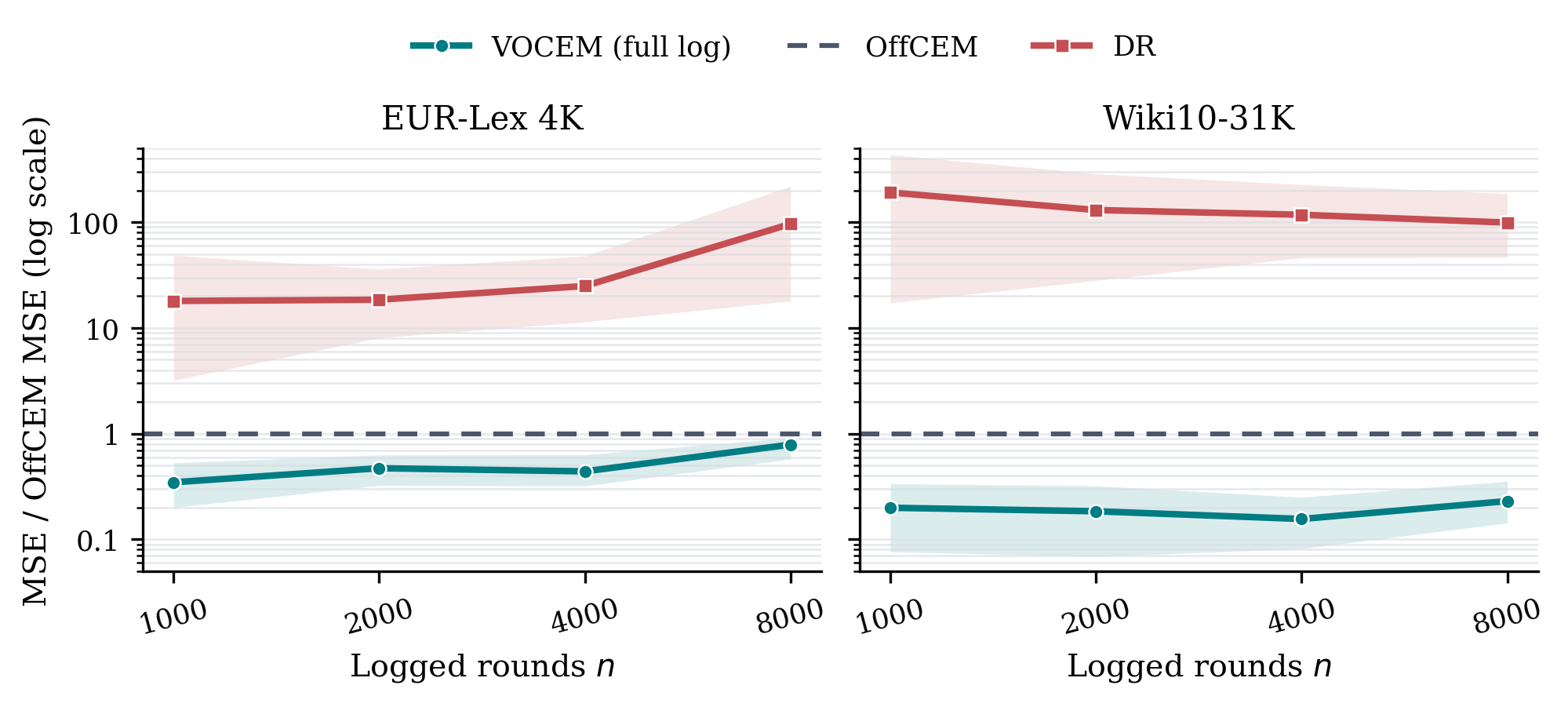}
  \caption{Real-data ablation with the true conditional reward mean.
  Points are MSE ratios to oracle-regression OffCEM; ribbons are pointwise
  paired-bootstrap 95\% confidence intervals over 100 repetitions.  The
  dashed line is OffCEM.}
  \label{fig:real-oracle-full-sample}
\end{figure}
\FloatBarrier

Figure~\ref{fig:real-oracle-full-sample} strengthens the attribution of the
main result.  VOCEM remains below OffCEM in all eight conditions, with every
upper confidence bound below one.  Its MSE ratio ranges from $0.347$ to
$0.788$ on EUR-Lex 4K and from $0.156$ to $0.230$ on Wiki10-31K.  In contrast,
the action-weighted DR endpoint has between $18.0$ and $191.2$ times the
corresponding OffCEM MSE.  Thus the same-log gains are not an artifact of the
particular learned reward model: they persist when reward-model error is
removed and only the correction weights differ.  As in the main experiments,
this ablation concerns a coefficient refitted on the evaluation log; it does
not establish that a coefficient can be transported to an independent log.

\subsection{Additional OPE baselines}
\label{app:additional-baselines}

We complement the comparisons in the main text with three standard OPE
baselines: inverse propensity scoring (IPS) and the direct method (DM).  These experiments
use exactly the same logs, policies, cluster assignments, reward predictions,
replications, and reporting protocol as the corresponding experiments in Section~\ref{sec:experiments}.

For completeness, the additional estimators are
\begin{align*}
    \Vhat_{\mathrm{IPS}}(\pi)
    =
    \frac{1}{n}\sum_{i=1}^n
    w_a(X_i,A_i)R_i,
    \qquad
    \Vhat_{\mathrm{DM}}(\pi)
    =
    \frac{1}{n}\sum_{i=1}^n
    \widehat q(X_i,\pi).
\end{align*}
IPS therefore uses the same action-level importance ratios as DR but applies
them directly to the observed rewards. DM uses only the learned reward regression and contains no
importance-weighted correction.

We report DM only in the real-data experiment with an estimated reward
model.  In the synthetic benchmark and in the real-data oracle-regression
ablation, all model-based estimators are supplied with
$\widehat q=q$ by construction.  Including DM there would therefore amount
to evaluating a direct estimator using the oracle conditional reward mean,
rather than a practically estimated reward model, and we omit it from those
comparisons.

\paragraph{Synthetic experiments.}
\begin{figure}[!tb]
    \centering
    \includegraphics[width=\linewidth]{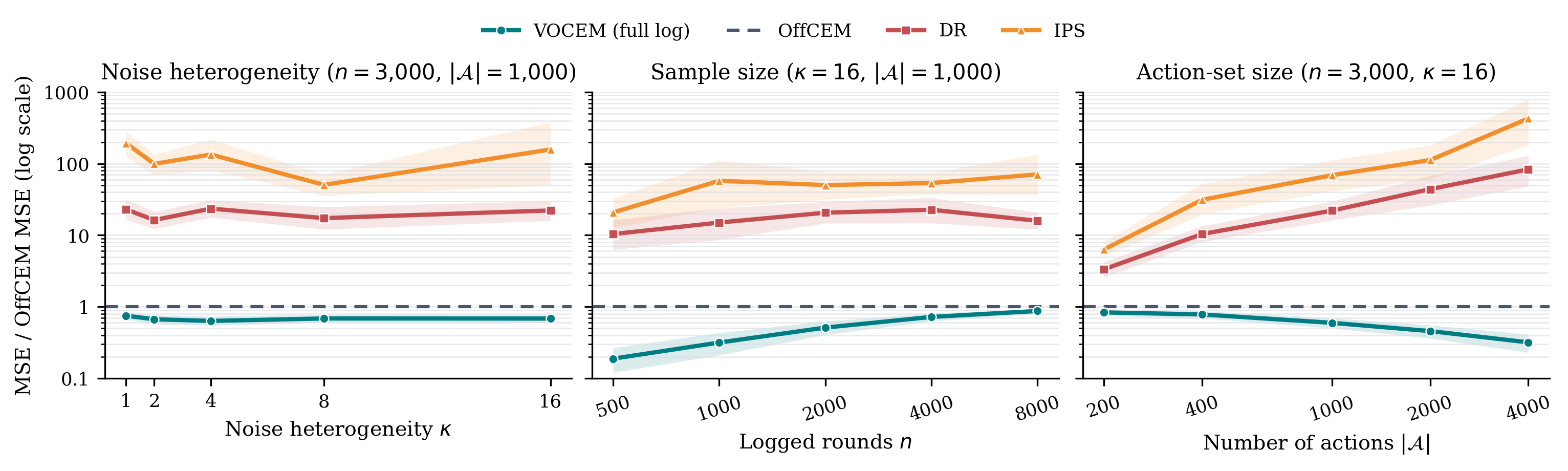}
    \caption{
    Additional baselines in the synthetic experiments.
    Points are MSE ratios to OffCEM and ribbons are pointwise
    paired-bootstrap 95\% confidence intervals over 300 repetitions.
    The dashed line is OffCEM.  IPS uses action-level importance
    weights; DM is omitted because the experiment supplies the oracle
    regression $\widehat q=q$ to the model-based estimators.
    }
    \label{fig:all-estimators-synthetic}
\end{figure}

Figure~\ref{fig:all-estimators-synthetic} repeats all three synthetic sweeps
with IPS included.  The additional baselines reinforce the
large-action motivation for cluster-level correction.  IPS is highly
variable throughout and deteriorates particularly sharply as the number of
actions increases.
DR is generally more stable than IPS because it importance-weights reward
residuals rather than raw rewards, but it likewise deteriorates as the
action space grows.  In contrast, the same-log VOCEM estimator remains
below the OffCEM reference throughout all three sweeps, as reported in the
main text.

\paragraph{Learned reward model.}
Figure~\ref{fig:all-estimators-real} adds IPS and DM to the
real-data experiments with five-fold out-of-fold reward regression.
Action-level IPS has substantially larger MSE than the cluster-based
estimators on both datasets, consistent with the weak practical overlap of
the constructed logging and target policies.  DR improves considerably over
IPS, demonstrating the benefit of applying the action-level weights to
reward residuals rather than directly to rewards, but it remains markedly
less stable than the estimators whose corrections avoid unrestricted
action-level weighting.

DM is competitive on Wiki10-31K, indicating that the
out-of-fold reward regression itself carries substantial predictive
information there. 

\begin{figure}[!tb]
    \centering
    \includegraphics[width=0.82\linewidth]{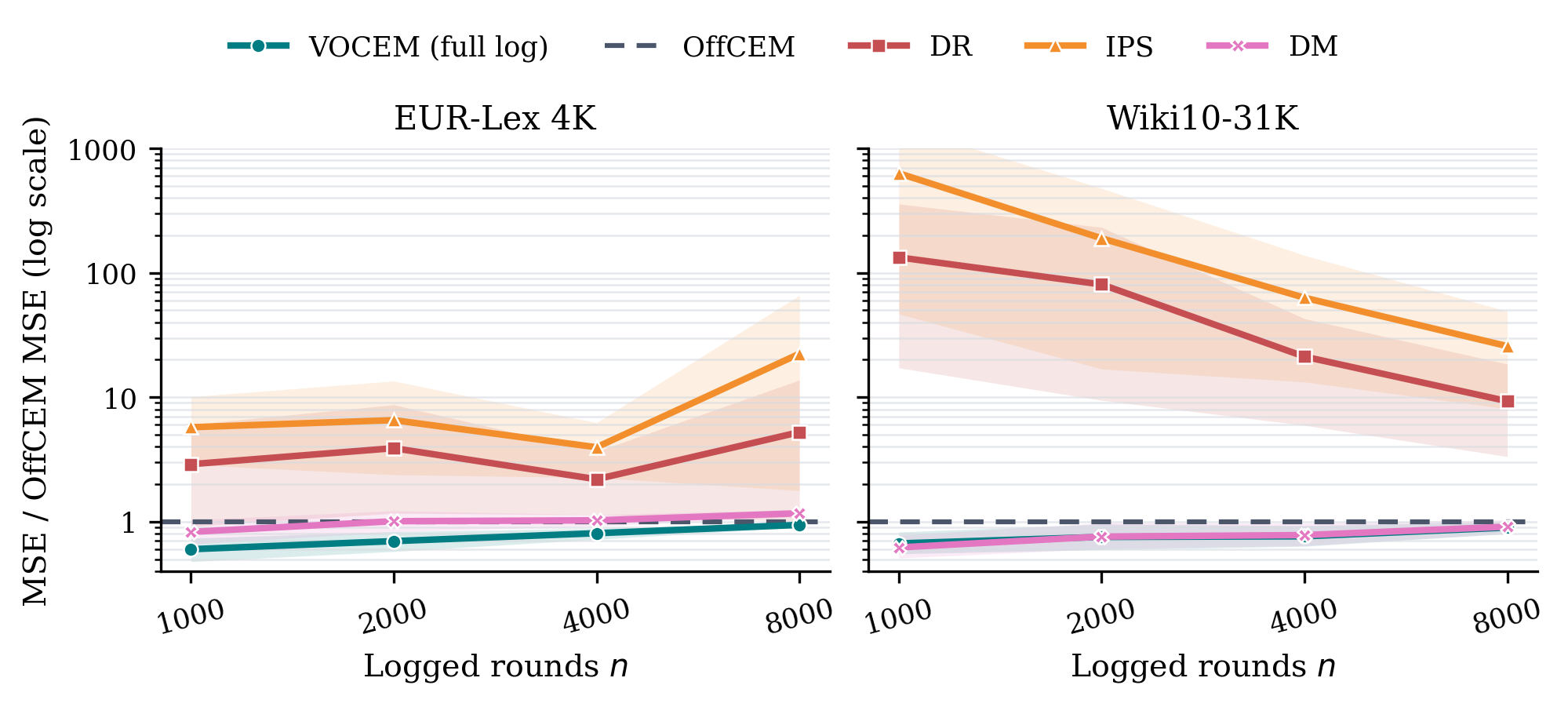}
    \caption{
    Additional baselines on EUR-Lex 4K and Wiki10-31K with the learned
    five-fold out-of-fold reward model.  Points are MSE ratios to OffCEM
    and ribbons are pointwise paired-bootstrap 95\% confidence intervals
    over 100 repetitions.  The dashed line is OffCEM.
    }
    \label{fig:all-estimators-real}
\end{figure}

\paragraph{Oracle-regression ablation.}
Finally, Figure~\ref{fig:all-estimators-real-oracle} augments the
oracle-regression ablation with IPS.  Once reward-model error is
removed, VOCEM remains substantially below OffCEM on both datasets, whereas
IPS and DR lie above the OffCEM reference.  IPS is especially
unstable.

The numerical scale in this figure should be interpreted relative to the
oracle-regression OffCEM denominator: supplying the true conditional mean
makes OffCEM itself much more accurate, so an estimator whose absolute error
changes little can have a much larger normalized MSE than in the
learned-regression experiment.  The comparison therefore further isolates
the role of the correction weights.  With regression error removed,
VOCEM's advantage over OffCEM persists, while direct action-level weighting
remains costly under the weak-overlap regime.

\begin{figure}[!htb]
    \centering
    \includegraphics[width=0.82\linewidth]{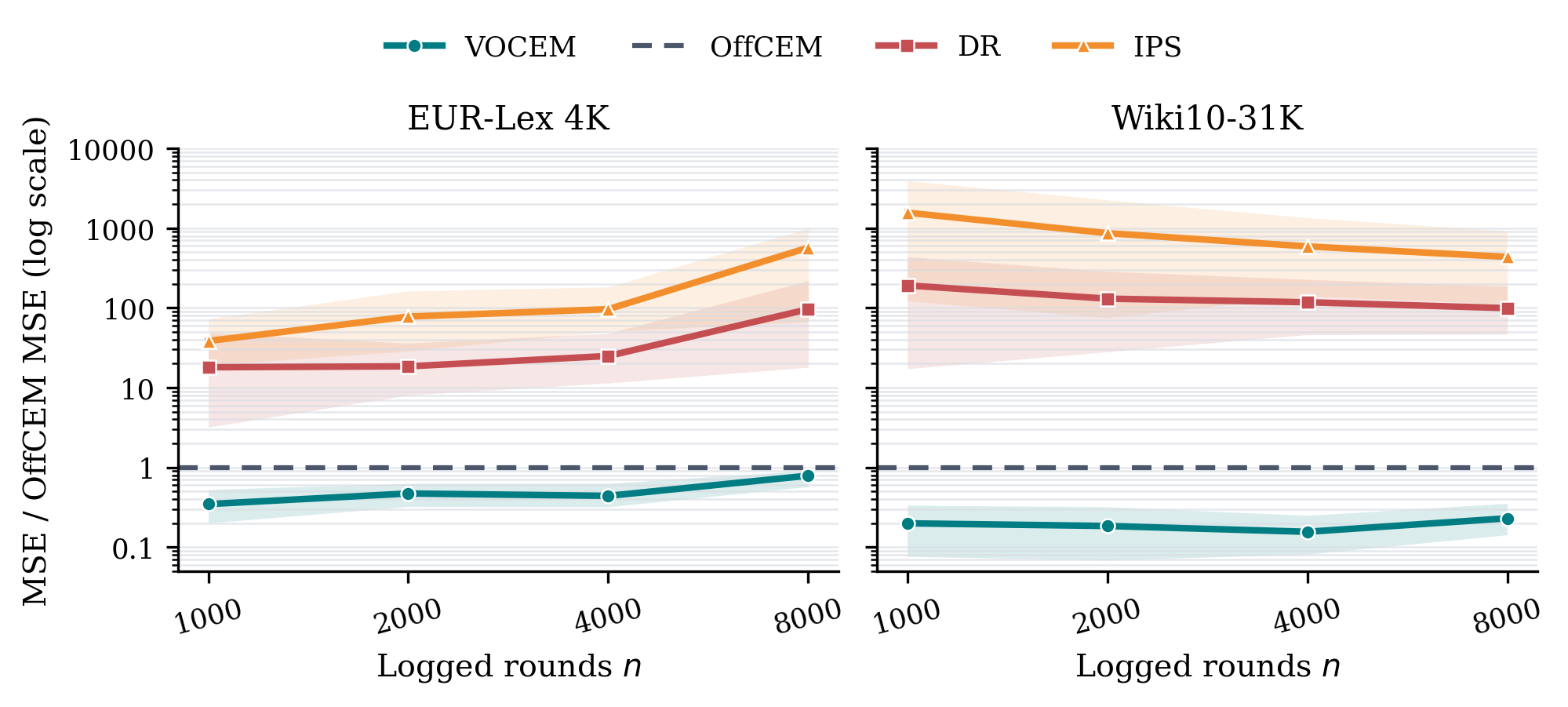}
    \caption{
    Additional baselines for the real-data oracle-regression ablation.
    All model-based estimators receive the true conditional mean
    $\widehat q=q$.  Points are MSE ratios to oracle-regression OffCEM and
    ribbons are pointwise paired-bootstrap 95\% confidence intervals over
    100 repetitions.  The dashed line is OffCEM.  DM is omitted because
    using it here would amount to direct evaluation with the oracle
    conditional reward model.
    }
    \label{fig:all-estimators-real-oracle}
\end{figure}

\section{Synthetic Experiment under Deficient Support}
\label{app:deficient-support}

We complement the full-support experiment by progressively removing actions
from the support of the logging policy. All data-generating parameters are
kept at their default values, with \(n=3{,}000\),
\(\lvert\mathcal A\rvert=1{,}000\), and \(\kappa=16\). We consider deficient
fractions
\(
  \rho\in\{0,0.3,0.5,0.7,0.9\}.
\)
For each \(\rho\), we select a fixed subset
\(\mathcal A_b^{(\rho)}\subseteq\mathcal A\), shared across repetitions, and
define
\[
  \pi_b^{(\rho)}(a\mid x)
  =
  \frac{
    \pi_b(a\mid x)\mathbbm{1}\{a\in\mathcal A_b^{(\rho)}\}
  }{
    \sum_{a'\in\mathcal A_b^{(\rho)}}\pi_b(a'\mid x)
  }.
\]
The supports are nested as \(\rho\) increases, and at least one action is
retained in every cluster. Hence cluster support continues to hold even though
action support fails. The target policy is left unchanged. Each condition is
repeated 300 times.

We consider two reward-model specifications. In the first, we use the exact
model \(\widehat q=q\), so deficient support is the only difference from the
experiment described above. In the second, we introduce a controlled error,
\(
  \widehat q(x,a)=q(x,a)-\delta
\), with $\delta=0.75$.
This second specification makes the consequence of missing action support
explicit: OffCEM can correct the error using the maintained cluster support,
whereas action-level corrections cannot recover the target-policy mass assigned
to actions that are never observed. In both experiments, VOCEM estimates
\(\widehat\lambda\) from the complete evaluation log.

\begin{figure}[t]
    \centering

    \begin{subfigure}[t]{0.43\linewidth}
        \centering
        \includegraphics[width=\linewidth]{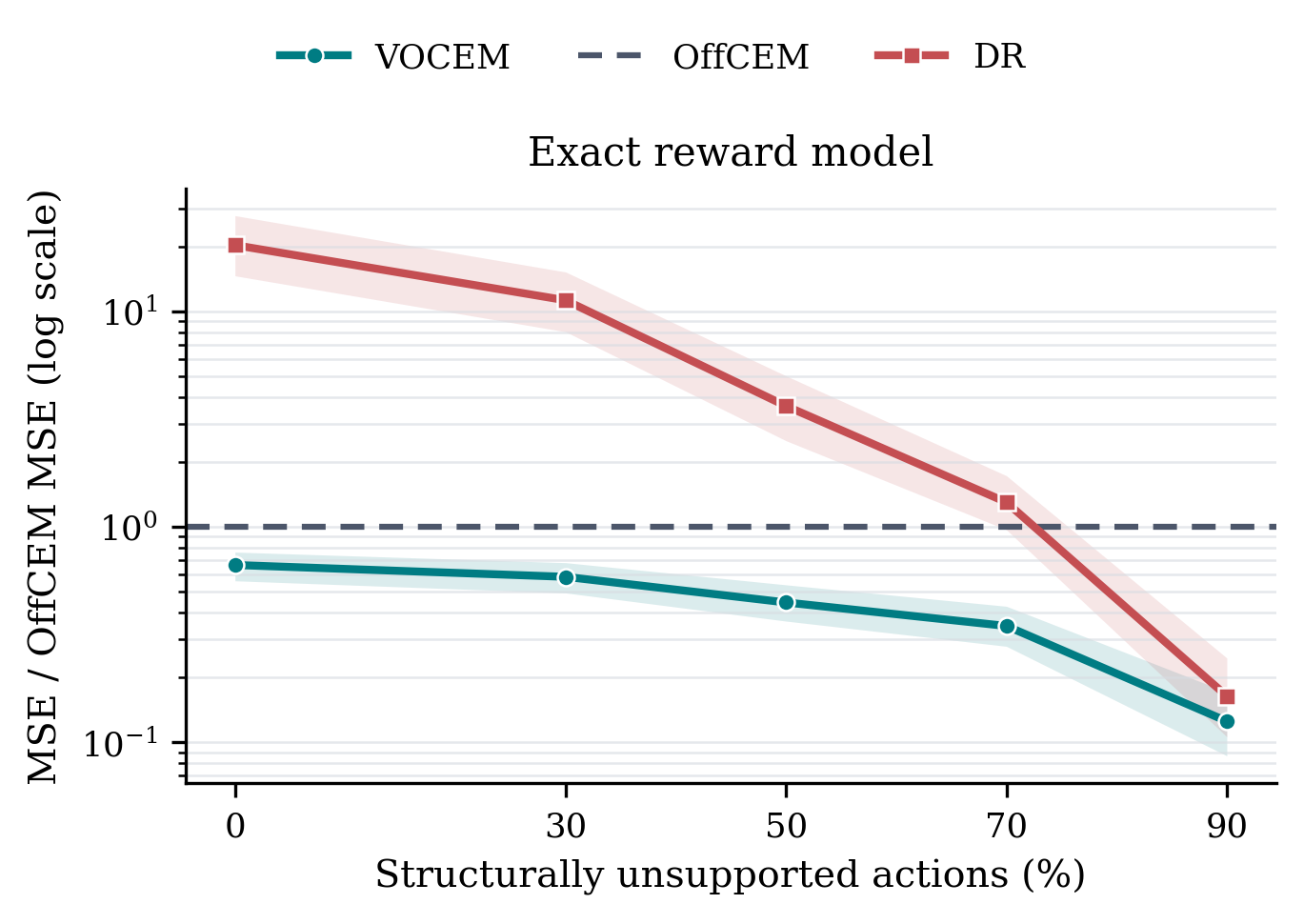}
        \caption{Exact reward model.}
        \label{fig:deficient-support-exact}
    \end{subfigure}
    \begin{subfigure}[t]{0.43\linewidth}
        \centering
        \includegraphics[width=\linewidth]{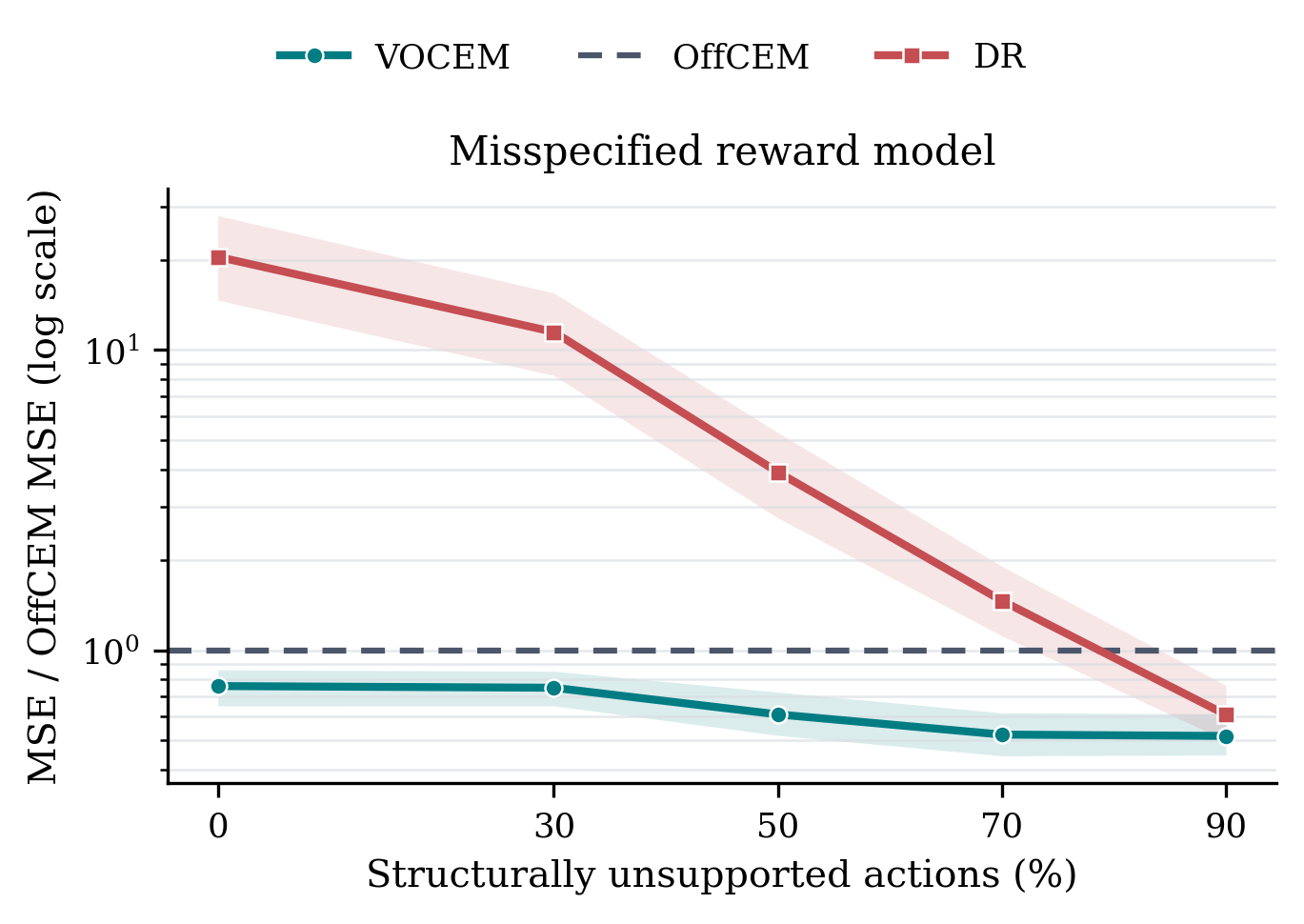}
        \caption{Misspecified reward model.}
        \label{fig:deficient-support-misspecified}
    \end{subfigure}

    \caption{
    MSE under deficient action support, normalized by the MSE of OffCEM.
    The fraction of structurally unsupported actions is progressively increased
    while preserving cluster-level support.
    Left: exact reward model.
    Right: misspecified reward model.
    Shaded regions indicate variability across Monte Carlo repetitions.
    }
    \label{fig:deficient-support}
\end{figure}

\paragraph{Results.}
Figure~\ref{fig:deficient-support} reports MSE relative to OffCEM. Under the
exact reward model, all estimators are approximately unbiased across support
levels, so the differences in MSE are driven almost entirely by variance.
As the action support of the logging policy contracts, the variance of DR drops
substantially because fewer action-level importance-weighted correction terms
remain active. Consequently, its relative MSE decreases from about \(20.3\)
under full support to well below that of OffCEM at \(\rho=0.9\). VOCEM
exhibits the same qualitative benefit while remaining substantially more stable:
its relative MSE decreases from approximately \(0.66\) at \(\rho=0\) to
\(0.13\) at \(\rho=0.9\). The estimated mixing coefficient correspondingly
moves toward the action-level correction, increasing from roughly \(0.34\) to
\(0.93\), as the variance penalty of that correction becomes progressively
smaller.

The misspecified reward model reveals the complementary bias effect. Since
cluster support is preserved, OffCEM remains approximately unbiased despite the
missing actions. DR, in contrast, cannot correct reward-model errors on actions
that are never observed, and therefore develops support-induced bias as the
target-policy mass on unsupported actions increases. VOCEM inherits this bias
only through the action-level component: it accepts progressively more of
the action-level bias in exchange for a reduction in variance.
Despite this trade-off, its MSE remains below that of OffCEM throughout the
experiment, with relative MSE decreasing from about \(0.76\) to \(0.52\).
Thus, in the considered misspecification regime, VOCEM is robust to deficient
action support because its low variance allows it to tolerate a moderate amount
of support-induced bias while still achieving a net reduction in MSE. This
robustness is therefore a bias--variance effect.

Overall, the two reward-model specifications separate the two effects of
deficient support. With an exact reward model, reduced support can lower MSE by
reducing the variance of action-level corrections. With reward-model error, the
same loss of support additionally introduces bias for estimators that rely on
those corrections. VOCEM balances these two effects: it can exploit the
variance reduction associated with action-level correction while tolerating
some of its induced bias, provided that the resulting variance gain is large
enough to compensate for the increase in squared bias.

\end{toappendix}




\section{Conclusion}
\label{sec:conclusion}

We introduced \vocem{} to improve off-policy evaluation when action-level
importance ratios induce high variance in DR estimation. By connecting
DR and OffCEM through a one-parameter family of residual corrections,
we make the choice between action-level and cluster-level weighting
an explicit variance-minimization problem. Under the same assumptions required by DR and OffCEM, every member of this family is
unbiased. We derive the population-optimal coefficient in closed form
and show that its estimator has variance no larger than either endpoint.
VOCEM estimates this coefficient directly from the logged data.

Across 15 synthetic and eight real-world benchmarks, VOCEM achieved
lower MSE than both DR and OffCEM, with gains observed under both oracle
and learned reward models. These results support adapting the residual
weights to the observed data as a way to improve estimation when DR's
action-level correction is unstable.

Future work could jointly learn the action clusters and interpolation
coefficient, adapting both the structure and strength of the residual
correction to improve estimation.

\clearpage
\subsection*{AI use statement}
Generative AI tools assisted with formulating theoretical claims through
iterative discussions and with developing and writing proofs. Generative AI tools also assisted with literature exploration,
interpreting results, and drafting and editing the manuscript. Coding
agents assisted with the Python implementation of the methods and
experiments, including synthetic-data generation based on the setup
of \citet{saito2023offcem}. The experimental design was developed by
the authors. Generative AI tools also assisted with interpreting results
and drafting and editing the manuscript. The authors checked the
mathematical claims and proofs, reviewed the code, and take responsibility
for the final content of this work, including all AI-assisted text,
claims, and code.

\bibliography{vocem_iclr2027}
\bibliographystyle{iclr2027_conference}

\end{document}